\documentclass[runningheads]{llncs}
\usepackage{amssymb}
\usepackage[numbers, square, sectionbib]{natbib}
\usepackage{color}
\usepackage[ruled,linesnumbered,boxed]{algorithm2e}
\usepackage{mathtools}
\usepackage{soul}
\usepackage{graphicx}
\SetKwInOut{Parameter}{Parameters}
\usepackage{url}
\urldef{\mailsa}\path|{alfred.hofmann, ursula.barth, ingrid.haas, frank.holzwarth,|
\urldef{\mailsb}\path|anna.kramer, leonie.kunz, christine.reiss, nicole.sator,|
\urldef{\mailsc}\path|erika.siebert-cole, peter.strasser, lncs}@springer.com|    
\newcommand{\keywords}[1]{\par\addvspace\baselineskip
\noindent\keywordname\enspace\ignorespaces#1}
\usepackage{graphicx}

\usepackage[table]{xcolor}
\usepackage{sistyle}
\SIthousandsep{,}

\usepackage{color}

\SetKwComment{Comment}{$\triangleright$\ }{}

\begin{document}
\title{Symbolic Graph Embedding using Frequent Pattern Mining}

\author{Bla\v{z} \v{S}krlj\inst{1,2} \and
Nada Lavra\v{c}\inst{2,1,3} \and Jan Kralj\inst{2}}
\authorrunning{\v{S}krlj et. al.}
%
\institute{Jo\v{z}ef Stefan International Postgraduate School \and
Jo\v{z}ef Stefan Institute, Slovenia 
\and
University of Nova Gorica, Slovenia
}

\maketitle 
\begin{abstract}
Relational data mining is becoming ubiquitous in many fields of study. It offers insights into behaviour of complex, real-world systems which cannot be modeled directly using propositional learning. We propose Symbolic Graph Embedding (SGE), an algorithm aimed to learn symbolic node representations. Built on the ideas from the field of inductive logic programming,  SGE first samples a given node's neighborhood and interprets it as a transaction database, which is used for frequent pattern mining to identify logical conjuncts of items that co-occur frequently in a given context. Such patterns are in this work used as features to represent individual nodes, yielding interpretable, symbolic node embeddings. The proposed SGE approach on a venue classification task outperforms shallow node embedding methods such as DeepWalk, and performs similarly to metapath2vec, a black-box representation learner that can exploit node and edge types in a given graph.
The proposed SGE approach performs especially well when small amounts of data are used for learning, scales to graphs with millions of nodes and edges, and can be run on an of-the-shelf laptop.
\end{abstract}

\section{Introduction}
Many contemporary databases are comprised of vast, linked and annotated data, which can be hard to exploit for various modeling purposes. In this work, we explore how learning from heterogeneous graphs (i.e. heterogeneous information networks with different types of nodes and edges) can be conducted using the ideas from the fields of symbolic relational learning and inductive logic programming  \cite{lavrac1994inductive}, as well as contemporary representation learning on graphs \cite{bengio2013representation}.

Relational datasets have been considered in machine learning since the early 1990s, where tools such as Aleph \cite{srinivasan2001aleph} have been widely used for relational data analysis.
However, recent advancements in deep learning, a field of subsymbolic machine learning, which allows for learning from relational data in the form of graphs, was shown as useful for many contemporary relational learning tasks at scale, including recommendation, anomaly detection and similar \cite{shi2018heterogeneous}. The state-of-the-art methodology exploits the notion of \emph{node embeddings}---nodes, represented using real-valued vectors. As such, node embeddings can be simply used with propositional learners such as e.g., logistic regression or neural networks. The node embeddings, however, directly offer little to no insight into connectivity patterns relevant for representing individual nodes.

In this work we demonstrate that symbolic pattern mining can be used for learning \emph{symbolic node embeddings} in heterogeneous information graphs. The main contributions of this work include:
\begin{enumerate}
\item An efficient graph sampler which samples based on a distribution of lengths of random walks, implemented in Numba, offering 15x faster sampling than a Python-native implementation, scaling to graphs with millions of nodes and edges on an of-the-shelf laptop.
\item Symbolic graph embedding (SGE), a symbolic representation learner that is explainable 
and achieves state-of-the-art performance for the task of node classification.
\item Evidence that symbolic node embeddings can perform comparably to black-box node embeddings, whilst requiring \emph{less} space and data.
\end{enumerate}

This paper is structured as follows. We first discuss the related work (Section~\ref{sec:rel}), followed by the description of the proposed approach (Section~\ref{sec:gar}), its computational and spatial complexity (Section~\ref{sec:complexity}), and its empirical evaluation (Section~\ref{sec:emp}). We finally discuss the obtained results and potential further work (Section~\ref{sec:fw}).

\section{Related work}
\label{sec:rel}


Symbolic representation learning has already been considered in the early 1990s in the inductive learning community, when addressing multi-relational learning problems through the so-called  \emph{propositionalization} approach  \cite{lavrac1994inductive}. The goal of propositionalization is to transform multi-relational data into real-valued vectors describing the individual training instances, that are a part of a relational data structure. The values of the vectors are obtained by evaluating a relational feature (e.g., a conjunct of conditions) as true (value 1) or false (value 0). For example, if all conditions of a conjunct are true, the relational feature is evaluated as true, resulting in value 1, and gets value 0 otherwise. 
We next discuss the approaches which were most influential for this work.
The in-house developed Wordification \cite{perovvsek2015wordification} explores how relational databases can be unfolded into bags of relational words, used in the same manner as done in the area of natural language processing via Bag-of-words-based representations. Wordification, albeit very fast, can be spatially expensive, and was designed for SQL-based datasets. Our work was also inspired by the recently introduced HINMINE methodology \cite{Kralj2018}, where Personalized PageRank vectors were used as the propositionalization mechanism. Here, each node is described via its probability to visit any other node, thus, a node of a netwrk is described using a distribution over the remainder of the nodes. Further, propositionalization has recently been explored in combination with artificial neural networks \cite{francca2014fast}, and as a building block of deep relational machines \cite{dash2018large}.

Frequent pattern mining is widely used for identifying interesting patterns in real-world transaction databases. Extension of this paradigm to graphs was already explored  \cite{inokuchi2000apriori}, using the Apriori algorithm \cite{agrawal1994fast} for the pattern mining. In this work we rely on the efficient FP-Growth \cite{borgelt2005implementation} algorithm, which employs more structured counting compared to Apriori using fp-trees as the data structure. Frequent pattern mining is commonly used to identify logical patterns which appear above a certain e.g., frequency threshold. Efficiently mining for such patterns remains a lively research area on its own, and can be scaled to large computing clusters \cite{han2007frequent}.

The proposed work also explores how a given graph can be sampled, as well as embedded efficiently. Many contemporary node representation learning methods, such as node2vec \cite{grover2016node2vec}, DeepWalk \cite{perozzi2014deepwalk}, PTE \cite{tang2015pte} and metapath2vec \cite{dong2017metapath2vec}, exploit such ideas in combination with e.g., the skip-gram model in order to obtain low-dimensional embeddings of nodes. Out of the aforementioned methods, only metapath2vec was adapted specifically to operate on \emph{heterogeneous information networks}, i.e. graphs with additional information on node and edge types. 
It samples pre-defined meta paths, yielding type-aware node representations which serve better for classifying e.g., different research venues to topics.

Heterogeneous (non-attributed) graphs are often formalized as RDF triplets. 
Relevant methods, which explore how such triplets can be embedded are considered in \cite{cochez2017global}, as well as in \cite{ristoski2016rdf2vec}. The latter introduced RDF2vec, a methodology for direct transformation of a RDF database to the space of real-valued entity embeddings.
Understanding how such graphs can be efficiently sampled, as well as embedded into low-dimensional, real-valued vectors is a challenging problem on its own.

\section{Proposed SGE algorithm}
\label{sec:gar}
\begin{figure}[t]
\centering
\includegraphics[width=.60\linewidth]{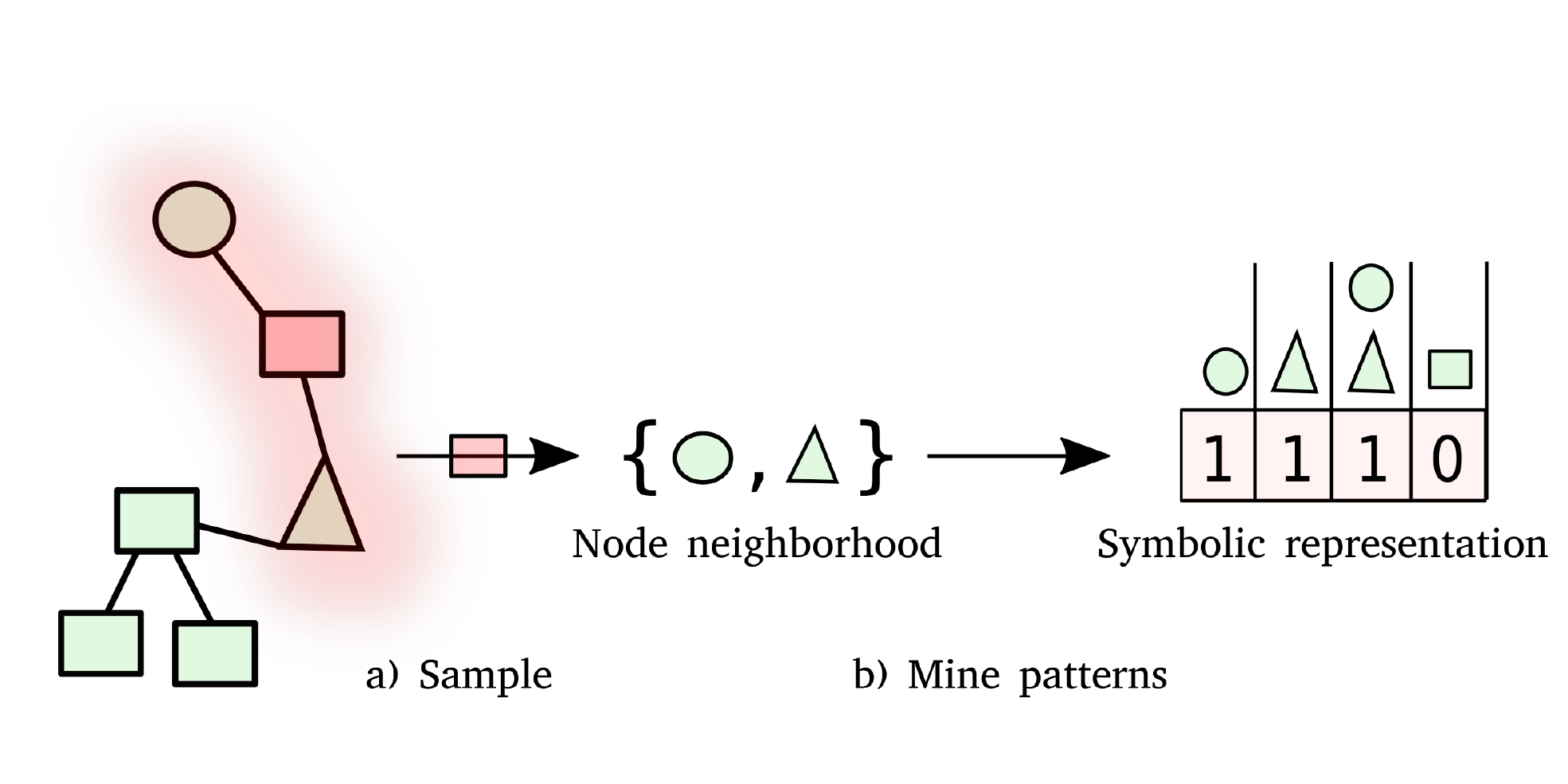}
\caption{Schematic represenatation of SGE. The red square's neighborhood (red highlight) is first used to construct symbolic features, forming a propositional graph representation (part a of the figure). The presence of various symbolic patterns (FP) is recorded and used to determine feature vectors for individual nodes (part b of the figure). 
The obtained representation can be used for subsequent data analysis tasks such as classification or visualization.}
\label{fig:scheme}
\end{figure}

In this section we describe Symbolic Graph Embedding (SGE), a new algorithm for symbolic node embedding. The algorithm is summarized in Figure~\ref{fig:scheme}. The algorithm consists of two basic steps. First, for each node in an input graph, the neighborhood of the node is sampled (Section~\ref{sec:sampling}). Next, the patterns, emerging from the walks around a given node are transformed into a set of features whose values describe the node (Section \ref{sec:prop}). In this section, we first introduce some basic definitions and then explain both steps of SGE in more detail. 

\subsection{Overview and definitions}
We first define the notions of a graph as used in this work.

\begin{definition}[Graph]
A \emph{graph} is a tuple $G = (N, E)$, where $N$ is a set of nodes and $E$ is a set of edges. The elements of $E$ can be either subset $N$ of size $2$ (e.g., $\{n_1, n_2\}\subseteq N$), in which case, we say the graph is \emph{undirected}. Alternatively, $E$ can consist of ordered pairs of elements from $N$ (e.g., $(n_1, n_2)\in N\times N$) -- in this case, the graph is \emph{directed}.
\end{definition}

In this work we focus on directed graphs, yet the proposed methodology can also be extended to undirected ones.
In this work we also use the notion of a \emph{walk}.

\begin{definition}[Walk]
Given a directed graph $G=(N, E)$, a \emph{walk} is any sequence of nodes $n_1, n_2\dots, n_k\in N$ so that each pair $n_i, n_{i+1}$ of consecutive nodes is connected by an edge, i.e. $(n_i, n_{i+k})\in E$.
\end{definition}

Finally, we define the notion of node embedding as used throughout this work.

\begin{definition}[Symbolic node embedding]
Given a directed graph $G=(N, E)$, a $d$-dimensional node embedding of graph $G$ is a matrix $\mathcal{M}$ in a vector space $\mathbb{R}^{|N| \times d}$, i.e. $\mathcal{M} \in \mathbb{R}^{|N| \times d}$. Such embedding is considered \emph{symbolic}, when each column represents a symbolic expression, which, when evaluated against a given node's neighborhood information, returns a real number representing a given node.
\end{definition}

We first discuss the proposed neighborhood sampling routine, followed by the description of pattern learning as used in this work.
\subsection{Sampling node neighborhoods}
\label{sec:sampling}

Sampling a given node's local and global neighborhoods offers insights into connectivity patterns of the node with respect to the rest of the graph. Many contemporary methods resort to node neighborhood sampling for obtaining the node co-occurrence information. In this work we propose a simple sampling scheme which produces a series of graph walks. The walks can be further used for learning tasks---in this work, we use them to produce node representations.  Building on recent research ideas \cite{leskovec2006sampling,maiya2010sampling}, the proposed scheme consists of two steps: selection of walk sampling distribution and sampling.
We first discuss the notion of distribution-based sampling, followed by the implemented sampling scheme.

\subsubsection{Distribution-based sampling}
Our algorithm is based on the assumption that when learning a representation of a given node, nodes at various distances from the considered node are relevant. In order to use the information on neighborhood nodes, we sample several random walks starting at each node of the graph. Because real-world graphs are diverse, it is unlikely that the same sampling scheme would suffice for arbitrary graphs. To account for such uncertainty, we introduce the notion of \emph{walk distribution vector}, a vector describing how many walks of a certain length shall be sampled. Let $w \in \mathbb{R}^{s}$ denote a vector of length $s$ (a parameter of the approach). The $i$-th value of the vector corresponds to the proportion of walks of length $i$ that are to be sampled. Note that the longest walk that can occur is of length $s$. For example consider the following vector $w$ of length $s = 4$,
$w = [0.2,0,0.5,0.3].$
Assuming we sample e.g, 100 random walks, 20 walks will be of length one, zero of length 2, 50 of length three and 30 of length four. As $w$ represents a probability distribution of different walk lengths to be sampled, $\sum_{i=1}^{s}w_i = 1$ must hold.

Having defined the formalism for describing the number of walks of different lengths, we have yet to describe the following two aspects in order to fully formalize the proposed sampling scheme:
how to parametrize $w$ and walk efficiently?
\begin{algorithm}[b]
\KwData{A graph $G = (N,E)$}
\Parameter{Starting node $n_i$, walk length $s$}
$c \leftarrow n_i$\; 
$\delta \leftarrow$ 0\;
$\mathcal{W} \leftarrow$  multiset\;
\For{$\alpha \in [1 \dots s]$ }
{
        $o$ := Uniform($N_G(c)$) \Comment*[r]{Select a random node.}
        $\mathcal{W} \leftarrow \mathcal{W} \cup (o, \alpha)$\Comment*[r]{Store visited node.}
        $c \leftarrow o$\;
}

\KwResult{A random walk $\mathcal{W}$}
\caption{Order-aware random walker.}
 \label{algo:rw}
\end{algorithm}

\subsubsection{How to parametrize $w$}
\label{sec:par}
We next discuss the considered initialization of the probability vector $w$. We attempt to model such vector by assuming a prior \emph{walk length} distribution, from which we first sample $\phi$ samples---these samples represent different random walk \emph{lengths}. In this work, we consider Uniform walk length distribution, where the $i$-th element of vector $w$ is defined as $\frac{1}{s}$, where $s$ represents the length of $w$ (maximum walk length).

The considered  variant of graph sampling procedure does not take into account node or edge types. One of the purposes of this work is to explore whether such na\"ive sampling---when combined with symbolic learning---achieves good performance. The rationale for not exploring how to incorporate node and edge types is thus twofold: First, we explore whether symbolic learning, as discussed in the next section, detects heterogeneous node patterns on its own, as the node representations are discrete and could, as such, provide such information. 
Further, as exact node information is kept intact, and each node can be mapped to its type, the node types are implicitly incorporated.
Next, we believe that by selecting the appropriate prior walk distribution $w$, node types can be to some extent taken into account (yet this claim depends largely on a given graph's topology).

\subsubsection{How to walk efficiently}
An example random walker, which produces walks of a given length (used in this work) is formalized in Algorithm~\ref{algo:rw}.
Here, we denote with $N_G(n_i)$ the neighbors of the $i$-th node. The Uniform($N_G(c)$) represents a randomly picked neighbor of a given node $c$, where picking each neighbor is equiprobable. We mark such picked node with $o$. Note that the walker is essentially a probabilistic depth-first search.
The algorithm returns a list of tuples where each tuple $(o, \alpha)$ contains both the visited node $o$ and the step $\alpha$ at which the node was visited. 
On line 6, we append to a current walk a tuple, comprised of a certain node and the overall walk length, it is a part of. Note that such inclusion of node IDs is suitable in a learning setting, where e.g., part of the graph's labels are not known and are to be predicted using the remainder of the graph. Even though inclusion of such information might seem redundant, we would like to remind the reader that the presented algorithm represents only a single random walker for a single walk length. In reality, multiple walkers yielding walks of different lengths are simulated, since including such positional (walk length) information can be beneficial for the subsequent representation learning step.
The proposed Algorithm~\ref{algo:rw} represents a simple random walker. In practice, thousands of random walks are considered. As discussed, their lengths are distributed according to $w$. In theory, one could learn the optimal $w$ by using e.g., stochastic optimization, yet we explore a different, computationally more feasible approach for obtaining a given $w$.
We next discuss the notion of symbolic pattern mining and final formalization of the proposed SGE algorithm.

\subsection{Symbolic pattern mining}
\label{sec:prop}
In the previous section we discussed how a given node's neighborhood can be efficiently sampled. In this section we first discuss the general idea behind forming node representations, followed by a description of the frequent pattern mining algorithms employed.

\subsubsection{Forming node representations}


Algorithm~\ref{algo:rw} outputs a multiset comprised of nodes, represented by (node id, walk order) tuples. Such multisets are in the following discussion considered as \emph{itemsets}, as this is the terminology used in \cite{borgelt2005implementation}.
In the next step of SGE, we use the itemsets to obtain individual node representations. We first give an outline of this step, and provide additional details in the next section.
The set of all nodes is considered as a transaction database, and the itemsets comprised of node id and walk order are used to identify frequent patterns (of tuples). Best patterns, selected based on their frequency of occurrence, are used as \emph{features}. The way of determining the best patterns is approach specific, and is discussed in the following paragraphs.
Feature values are determined based on the pattern identification method, and are either real-, natural- or binary-valued. Intuitively, they represent the presence of a given node pattern in a given node's neighborhood. 


\subsubsection{Frequent pattern mining}
\label{sec:symba}

We next discuss the frequent pattern mining approaches explored as part of SGE. The described approaches constitute the \emph{findPatterns} method discussed in the next section.
For each node, a multiset of (node ID, walk length) tuples is obtained. In each of the described approaches, the result is a transformation of the set of multisets, describing the network nodes, into a set of feature vectors describing these nodes.

\begin{description}
\item \textbf{Relational BoW}. This paradigm leverages the Bag-of-words (BOW) constructors widely known in natural language processing \cite{zhang2010understanding}.
Here, the tuples forming the itemsets, output by Algorithm~\ref{algo:rw}, are considered as words. Thus, each word is comprised of a node and the order of a random walk in which that node was identified as connected to the node for which the representation is being constructed.

For the purpose of BOW construction, we consider the multiset of (node ID, walk length) tuples, generated by random walks that start at node $n$. This multiset is viewed as a ``\emph{node document}'', consisting of individual words---i.e. the tuples contained in the multiset. The number of total features, $d$, is a parameter of the SGE algorithm. We consider the following variations of this paradigm for transforming each node ``document'' $\mathfrak{T}_n$ into one feature vector of size $d$:

\begin{itemize}
\item \textbf{Binary}. In a binary conversion, the values of the vector represent the presence or absence of a given tuple $k$-gram (a combination of $k$ tuples; $k$ is a free parameter of SGE) in the set of random walks associated with a given node document. The features, represented by such tuple $k$-grams, can have values of either 0 or 1.
\item \textbf{TF}. Here, counts of a given $k$-gram $t$ in a given node document $\mathfrak{T}_n$ are used as feature values ($TF_{t,\mathfrak{T}_n}$). The values are integers. Note that $TF_{t,\mathfrak{T}_n}$ represents the multiplicity of a given tuple $k$-gram in the multiset (node document).
\item \textbf{TF-IDF}. Here, TF-IDF weighting scheme is employed to weight the values of individual features. The obtained values are real numbers. Given a tuple $k$-gram $t$ and the transaction database $\mathfrak{T}$, it is computed as: 
\begin{align*}
\mbox{TF-IDF}(t,n) = (1 +\log \mbox{TF}_{t,\mathfrak{T}_n}) \cdot \log \frac{|N|}{\mathfrak{T}_t},
\end{align*}
where $TF_{t,\mathfrak{T}_N}$ is the number of $t$'s occurrences in a given document of node $n$ and $\mathfrak{T}_t$ is the overall occurrence of this $k$-gram in the whole transaction database.
\end{itemize}
\item \textbf{FP-Growth}.
This well known variant of association rule learning \cite{borgelt2005implementation} constructs a specialized data structure termed fp-tree, which is used to count combinations of tuples of different lengths. It is more efficient than the well known Apriori algorithm \cite{borgelt2003efficient,perego2001enhancing}.

For the purpuse of FP-growth, the obtained multiset $\mathfrak{T}$ is viewed as a set of \emph{itemsets} - a transaction database. For each itemset, only the set of unique tuples is considered as the input while their multiplicity is ignored. The FP-Growth algorithm next considers such non-redundant transaction databse $\mathfrak{T}$ to identify frequent combinations of tuples, similarly to the TF and TF-IDF schemes described above. The free parameter we consider in this work is \emph{support}, which controls how frequent tuple combinations shall be considered. Similarly to TF and TF-IDF schemes, once the representative tuple combinations are obtained, they are considered as features, whose values are determined based on their presence in a given node document, and are binary (0 = not present, 1 = present). Note that some of these features may correspond to the features generated by TF-IDF, however, in the case of FP-growth, we allow sizes of tuple combinations to be arbitrary, rather than fixed to $k$. Also unlike the BOW approaches, the dimension of the constructed feature vectors constructed is not fixed but is controlled implicitly by varying the value of the \emph{support} parameter.
\end{description}

\subsubsection{SGE formulation}

\begin{algorithm}[t]
\KwData{A graph $G = (N,E$)}
\Parameter{Number of walk samples $\nu$, sampling distribution $\eta$, pattern finder $r$, embedding dimensionality $d$, starting node $n_i$}
\KwResult{Symbolic node embedding $M$}
$\tau$ := generateSamplingVector($\eta$, $\nu$)\;
 $\mathfrak{T} \leftarrow$ multiset\;
\For{$o \in \tau$}{
    $\alpha \leftarrow$ $o$'s index  \Comment*[r]{Walk length.}
    $\mathcal{D} \leftarrow \{\}$\;
    \For{$k \in [1 \dots o]$}{
        $\mathcal{D}$ $\leftarrow \mathcal{D}$ $\cup $ Walk($G$, $n_i$, $\alpha$) \Comment*[r]{Sample with Algorithm~\ref{algo:rw}.}
    }
    $\mathfrak{T} \leftarrow \mathfrak{T}$ $ \cup $ $\mathcal{D}$ \Comment*[r]{Update walk object.}
}
$\mathcal{P}$ $\leftarrow$ findPatterns($\mathfrak{T}$, $r$) \Comment*[r]{Find patterns.}
$\mathcal{M} \leftarrow$  representNodes($\mathfrak{T}$, $\mathcal{P}$, $r$, $d$) \Comment*[r]{Represent nodes.}
\Return{$\mathcal{M}$}\;
\caption{Symbolic graph embedding.}
 \label{algo:rep}
\end{algorithm}

The formulation of the whole approach is given in Algorithm~\ref{algo:rep}. Here, first the sampling vector $w$ is constructed. Next, the vector is traversed. The $i$-th component of vector $w$ represents the number of walks of length $i$ that will be simulated. For each component of $w$ cell, a series of random walks (lines 6-8) is simulated, which produces sequences of nodes that are used to fill a node-level walk container $\mathcal{D}$. Thus, $\mathcal{D}$, once filled, consists of $o$ sets representing individual random walks of length $\alpha$. The walks are added into a single multiset, prior to being stored into the global transaction structure $\mathfrak{T}$. Once $w$ is traversed, frequent patterns are found (line 11), where the transaction structure $\mathfrak{T}$ comprised of all node-level walks is used as the input.
The findPatterns method in line 11 can be any method that takes a transaction database as input, the considered ones are discussed in the following section.
The top most frequent $d$ patterns are used as features, and represent the columns (dimensions) of the final representation $\mathcal{M}$. 
Here, the representNodes method (line 12) fills the values according to the considered weighting scheme (part of $r$)\footnote{Note that this method takes as input random walk samples for \emph{all} nodes.}.

\section{Computational and spatial complexity}
\label{sec:complexity}

In this section we discuss the computational aspects of the proposed approach. We split this section into two main parts, where we first discuss the complexity of the sampling, followed by the pattern mining part.

The time complexity of the proposed sampling strategy depends on the number of simulated walks and the walk lengths. The complexity of a single walk is linear with respect to the length of the walk. If we define the average walk length as $\overline l$, and the number of all samples as $\nu$, the spatial complexity, required to store all walks amounts to $\mathcal{O}(|N| \cdot \nu \cdot \overline l)$. As the complexity of a single random walk is linear with respect to the length of the walk, the considered sampling's time complexity amounts to $\mathcal{O}(\nu \cdot \overline l)$ for a single node. The proposed approach is also linear with respect to the number of nodes both in space and time.

The computational complexity of pattern mining varies based on the algorithm used for this step. The considered FP-Growth's complexity is linear with respect to the number of transactions, whereas its spatial complexity is, due to efficient counting employed, similarly efficient and does not explode as for example with the Apriori family of algorithms. The result of the pattern mining step is a $|N| \times d$ matrix, where $d$ is the number of patterns considered as features. Compared to e.g., metapath2vec and other shallow graph embedding methods, which yield a dense matrix, this matrix is \emph{sparse}, and potentially requires orders of magnitude less space for the same $d$\footnote{In practice, however, larger dimensions are needed to represent the set of nodes well by using symbolic representations.}.
As storing large dense matrices can be spatially demanding, the proposed sparse feature representation requires less space, especially if high-dimensional embeddings are considered (the black-box methods commonly yield dense representation matrices). The difference arises especially for very large datasets, where dense node representations can become a spatial bottleneck. We observe that $\approx$ 10\% of elements are non-zero, indicating that storing the feature space as a sparse matrix results in smaller time complexity. Worst case spatial complexity of storing the embedding, however, is for both types of methods $\mathcal{O}(|N| \cdot d)$.

\section{Empirical evaluation}
\label{sec:emp}
In this section we present the evaluation setting, where we demonstrate the performance of the proposed Symbolic Graph Embedding approach. We follow closely the evaluation introduced by metapath2vec, where the representation is first obtained, and next used for the classification task, where logistic regression is used as a classifier of choice. 
We test the performance on a heterogeneous information graph, comprised of authors, papers and venues\footnote{Accessible at \url{https://ericdongyx.github.io/metapath2vec/m2v.html}}. The task is to classify venues into one of eight possible topics. The dataset was first used for evaluation of metapath2vec, hence we refer to the original results when comparing with the proposed approach.  The considered graph consists of \num{2766148} nodes and
\num{2503628} edges, where the 133 venues are to be classified into correct classes.
We compare SGE against previously reported performances \cite{dong2017metapath2vec} of DeepWalk \cite{perozzi2014deepwalk}, LINE \cite{tang2015line}, PTE \cite{tang2015pte}, metapath2vec and metapath2vec++ \cite{dong2017metapath2vec}. All methods are considered state-of-the-art for black-box node representation learning. The PTE and two variations of metapath2vec can take into account different (typed) paths during sampling.

We tested the following SGE variants.
For pattern learning, we varied the TF-IDF, BoW and TF-, as well as the FP-Growth methods. 
The parameter search space used to obtain the results was as follows. The number of features = [500,1000,1500,2000,3000], considered vectorizers = [``TF-IDF",``TF",``FP-growth",``Binary"], relation order (relevant for TF-based vectorizers---the highest $k$-gram order considered) = [2, 3, 4], walks of lengths = [2, 3, 5, 10], and number of walk samples = [1000, 10000] were considered.
The support parameter of the FP-Growth parameter was varied in the range [3, 5, 8].
The Uniform walk length distribution was used. In addition to the proposed graph sampling (FS), a simple breadth-first search (BFS) that explores neighborhood of order two was also tested. We report the best performing learners' scores based on the type of the vectorizer and the sampling distribution.
Ten repetitions of ten-fold, stratified cross validation is used, the resulting micro and macro F1 scores are averaged to obtain the final performance estimate. We report the performance of logistic regression classifier when varying the percentage of training data.

\subsection{Results}
\begin{figure}[ht]
\centering
\includegraphics[width=\linewidth,height=5cm]{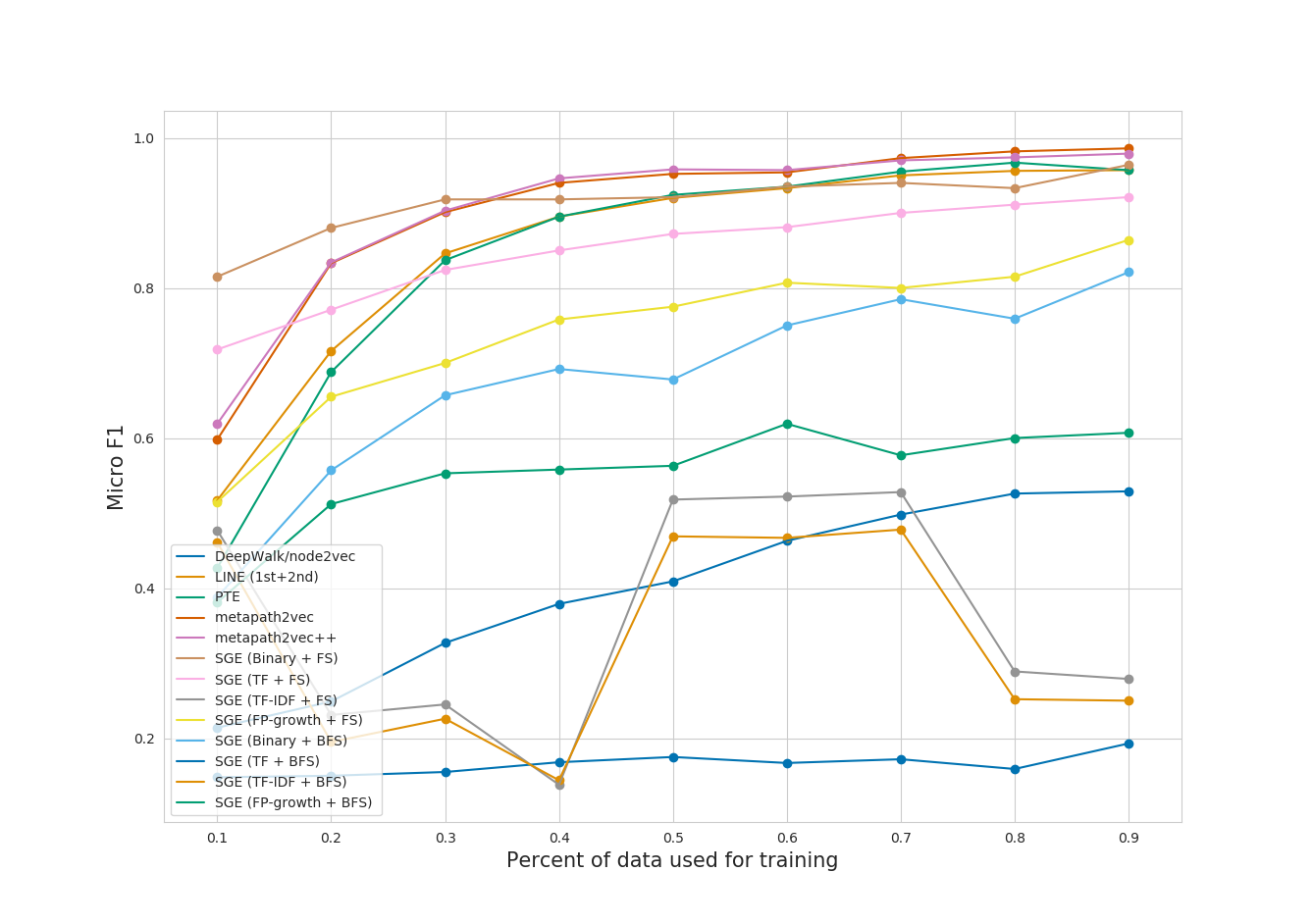}
\includegraphics[width=\linewidth,height=5cm]{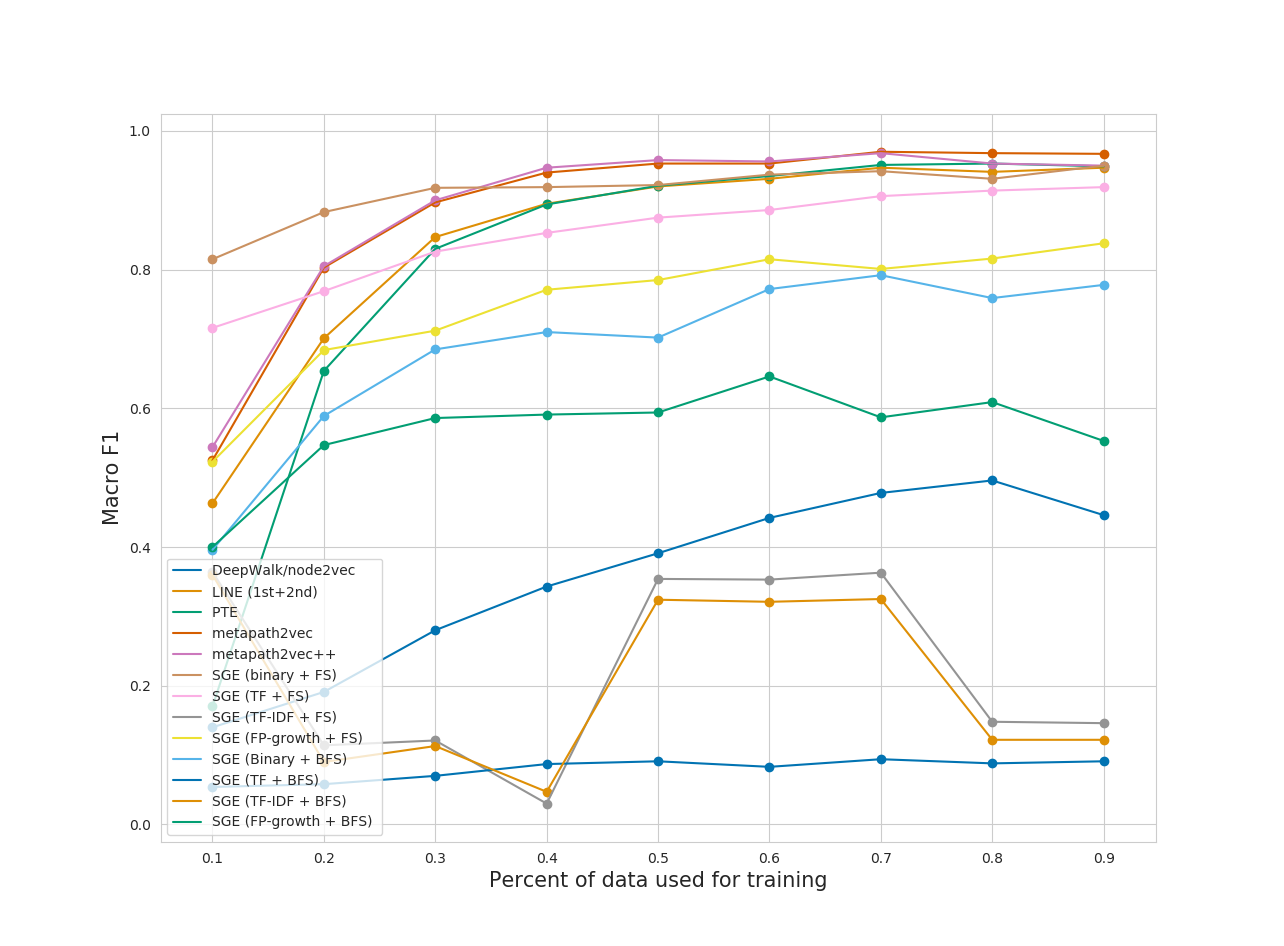}
\caption{Micro and macro F1 performance with respect to train percentages.}
\label{fig:percent}
\end{figure}

In this section we discuss in detail the results for the node classification task. The results in Table~\ref{tbl:res} are presented in terms of micro and macro F1 scores, with respect to training set percentage. We  visualize the performance of the compared representations  in Figure~\ref{fig:percent}.
\begin{table}[h]
\centering
\caption{Numeric results of the proposed SGE approach compared to the state-of-the-art approaches, presented in terms of micro and macro F1 scores, with respect to training set percentage. Best performing approaches are highlighted in green.}
\resizebox{0.70\textwidth}{!}{
\begin{tabular}{l|ccccccccccc}
\hline
Method / Percentage &  10\% & 20\% & 30\% & 40\% & 50\% & 60\% & 70\% & 80\% & 90\% \\ \hline
Macro-F1 &&&&&&&&&& \\ \hline
DeepWalk/node2vec  & 0.140 & 0.191 & 0.280 & 0.343 & 0.391 & 0.442 & 0.478 & 0.496 & 0.446\\
LINE (1st+2nd)  & 0.463 & 0.701 & 0.847 & 0.895 & 0.920 & 0.931 & 0.947 & 0.941 & 0.947\\
PTE & 0.170  & 0.654 & 0.830 & 0.894 & 0.921 & 0.935 & 0.951 & 0.953 & 0.949\\
 \rowcolor{green!8} metapath2vec  & 0.525 & 0.803 & 0.897 & 0.940 & 0.953 & 0.953 & \textbf{0.970} & \textbf{0.968} & \textbf{0.967}\\
 \rowcolor{green!8} metapath2vec++ & 0.544 & 0.805 & 0.900 & \textbf{0.947} & \textbf{0.958} & \textbf{0.956} & 0.968 & 0.953 & 0.950\\ \hline
 \rowcolor{green!8}  SGE (binary + FS) & \textbf{0.815} & \textbf{0.883} & \textbf{0.918} & 0.919 & 0.922 & 0.937 & 0.942 & 0.931 & 0.950\\
SGE (TF + FS) & 0.716 & 0.769 & 0.826 & 0.853 & 0.875 & 0.886 & 0.906 & 0.914 & 0.919\\
SGE (TF-IDF + FS) & 0.364 & 0.114 & 0.121 & 0.03 & 0.354 & 0.353 & 0.363 & 0.148 & 0.146\\
SGE (FP-growth + FS) &0.523&0.684&0.712&0.771&0.785&0.815&0.801&0.816&0.838 \\
SGE (Binary + BFS) & 0.396 & 0.589 & 0.685 & 0.710 & 0.702 & 0.772 & 0.792 & 0.759 & 0.778\\
SGE (TF + BFS) & 0.054 & 0.058 & 0.070 & 0.087 & 0.091 & 0.083 & 0.094 & 0.088 & 0.091\\
SGE (TF-IDF + BFS) & 0.360 & 0.090 & 0.113 & 0.047 & 0.324 & 0.321 & 0.325 & 0.122 & 0.122\\
SGE (FP-growth + BFS) & 0.400&0.547&0.586&0.591&0.594&0.646&0.587&0.609&0.553 \\
\hline
Micro-F1 &&&&&&&&&&& \\ \hline
DeepWalk/node2vec & 0.214 & 0.249 & 0.327 & 0.379 & 0.409 & 0.463 & 0.498 & 0.526 & 0.529\\
LINE (1st+2nd) & 0.517 & 0.716 & 0.846 & 0.895 & 0.920 & 0.933 & 0.950 & 0.956 & 0.957\\
PTE  & 0.427 & 0.688 & 0.837 & 0.895 & 0.924 & 0.935 & 0.955 & 0.967 & 0.957\\
 \rowcolor{green!8} metapath2vec & 0.598 & 0.833 & 0.901 & 0.940 & 0.952 & 0.954 & \textbf{0.973} & \textbf{0.982} & \textbf{0.986}\\
 \rowcolor{green!8} metapath2vec++  & 0.619 & 0.834 & 0.903 & \textbf{0.946} & \textbf{0.958} & \textbf{0.957} & 0.970 & 0.974 & 0.979\\ \hline
\rowcolor{green!8} SGE (Binary + FS) & \textbf{0.815} & \textbf{0.880} & \textbf{0.918} & 0.918 & 0.921 & 0.935 & 0.940 & 0.933 & 0.964\\
SGE (TF + FS) & 0.718 & 0.771 & 0.824 & 0.850 & 0.872 & 0.881 & 0.900 & 0.911 & 0.921\\
SGE (TF-IDF + FS) & 0.477 & 0.231 & 0.245 & 0.138 & 0.518 & 0.522 & 0.528 & 0.289 & 0.279\\
SGE (FP-growth + FS) &0.515&0.655&0.700&0.758&0.775&0.807&0.800&0.815&0.864 \\
SGE (Binary + BFS) & 0.388 & 0.557 & 0.657 & 0.692 & 0.678 & 0.750 & 0.785 & 0.759 & 0.821\\
SGE (TF + BFS) & 0.148 & 0.150 & 0.155 & 0.168 & 0.175 & 0.167 & 0.172 & 0.159 & 0.193\\
SGE (TF-IDF + BFS) & 0.461 & 0.195 & 0.226 & 0.144 & 0.469 & 0.467 & 0.478 & 0.252 & 0.250\\
SGE (FP-growth + BFS) & 0.381&0.512&0.553&0.558&0.563&0.619&0.577&0.600&0.607 \\
\hline
\end{tabular}
}
\label{tbl:res}
\end{table}

\begin{figure}[ht]
\centering
\includegraphics[width=0.5\linewidth]{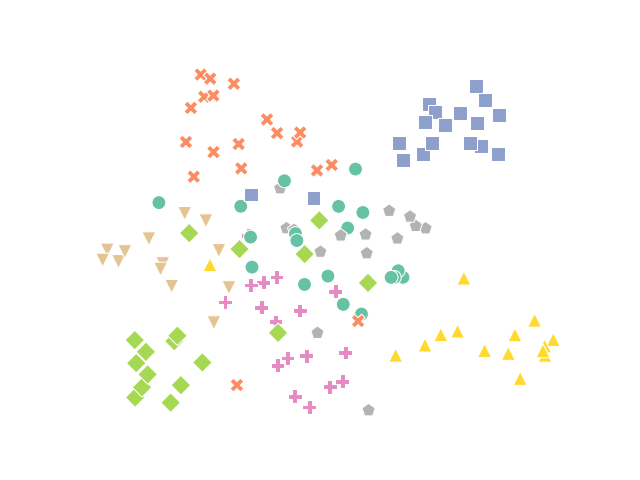}
\caption{UMAP projection of the best performing SGE embedding into 2D. Colors represent different types of venues (the class to be predicted). It can be observed that the obtained embeddings maintain the class-dependent structure, even though they were constructed in a completely unsupervised manner. The visualization was obtained using UMAP's default parameters.}
\label{fig:umap}
\end{figure}

The first observation is that shallow node embedding methods, e.g., node2vec and LINE,  do not perform as well as the best performing SGE variants (Binary with Uniform sampling). Further, we can observe that best performing SGE also outperforms metapath2vec and metapath2vec++, indicating that symbolic representations can (at least for this particular dataset) offer sufficient node description. The best performing SGE variant was the simplest one, with simple binary features obtained via fast sampling. Here, 10{,}000 walks were sampled and feature matrix of dimension 3000 was considered along with up to three-gram patterns.
Finally, we visualized the embeddings by projecting them to 2D using the UMAP algorithm \cite{mcinnes2018umap-software}. The resulting visualization, shown in Figure~\ref{fig:umap}, shows that the obtained symbolic node embeddings maintain the class structure of the data. 

\subsection{Implementation details and reproducibility}
\label{sec:imp}
In this section we discuss the details of the proposed SGE. The main part of the implementation is Python-based, where Numpy \cite{walt2011numpy} and Scipy \cite{jones2014scipy} libraries were used for efficient processing. The Py3plex library\footnote{\url{https://github.com/SkBlaz/Py3plex}} was used to parse the heterogeneous graph used as input \cite{vskrlj2018py3plex}. The final graph was returned as a MultiDiGraph object compatible with NetworkX \cite{osti_960616}. The TF, TF-IDF and Binary vectorizers implementations from the Scikit-learn library \cite{pedregosa2011scikit} were used. As the main bottleneck we recognized the graph sampling, which we further implemented using the Numba \cite{lam2015numba} framework for production of compiled code from native Python. After re-implementing the walk sampling part in Numba, we achieved approximately 15x speedup, which was enough to consider up to 10{,}000 walk samples of the large benchmark graph used in this work\footnote{The code repository is available at \url{https://github.com/SkBlaz/SGE}}.

\section{Discussion and conclusions}
\label{sec:fw}

In this work we compared the down-stream learning performance of symbolic features, obtained by sampling a given node's neighborhood, to the performance of black-box learners. Testing the approaches on the venue classification task, we find that Symbolic Graph Embedding offers similar performance on a large, real-world graph comprised of millions of nodes and edges. The proposed method outperforms the state-of-the-art shallow embeddings by up to $\approx$65\%, and heterogeneous graph embeddings by up to $\approx$27\% when only small percentages of the representation are used for learning (e.g., 10\%). The method performs comparably to metapath2vec and metapath2vec++ when the whole embedding is considered for learning. 
One of the most apparent results is the well performing Binary + Uniform SGE, which indicates that simply checking the presence of relational features potentially offers enough descriptive power for successful classification. This result indicates that certain graph patterns emerge as important, where their presence or absence in a given node's neighborhood can serve as relevant for classification. The TF-IDF-based SGE variants performed the worst, indicating that more complex  weighting schemes are not as applicable as in the other areas of text mining.
We believe the proposed methodology could be further compared with RDF2vec and similar triplet embedding methods.
The obtained symbolic embeddings were also explored qualitatively, where UMAP projection to 2D was leveraged to inspect whether the SGE symbolic node representations group according to their assigned classes. Such grouping indicates potential quality of the embedding, as venues of similar topics should be clustered together in the latent space.

\section{Acknowledgements}
We acknowledge the financial support from the Slovenian Research Agency through core research programmes P2-0103 and P6-0411 and project \emph{Semantic Data Mining for Linked Open Data} (financed under the ERC Complementary Scheme, N2-0078). The authors have received funding also from the European Union’s Horizon 2020 research and innovation programme under grant agreement No 825153 (EMBEDDIA). The work of the second author was funded by the Slovenian Research Agency through a young researcher grant (BS). We would finally like to thank to Jan Kralj for his insightful comments on formulation of the proposed framework and mathematical proofreading.

\bibliographystyle{splncs04}
\bibliography{references}

\end{document}